\documentclass[sigconf,table]{acmart}
\usepackage{tcolorbox}
\tcbuselibrary{skins}
\usepackage[nameinlink]{cleveref}
\usepackage{pbox}

\usepackage{tikz}
\usetikzlibrary{shapes.geometric, arrows}
\tikzstyle{startstop} = [rectangle, rounded corners, minimum width=3cm, minimum height=0.5cm,text centered, draw=black, fill=red!30]
\tikzstyle{io} = [trapezium, trapezium left angle=70, trapezium right angle=110, minimum width=1.5cm, minimum height=0.5cm, text centered, draw=black, fill=blue!30]
\tikzstyle{process} = [rectangle, minimum width=1.5cm, minimum height=0.5cm, text centered, draw=black, fill=orange!30]
\tikzstyle{decision} = [diamond, minimum width=1.5cm, minimum height=0.5cm, text centered, draw=black, fill=green!30]
\tikzstyle{arrow} = [thick,->,>=stealth]

\AtBeginDocument{%
  \providecommand\BibTeX{{%
    \normalfont B\kern-0.5em{\scshape i\kern-0.25em b}\kern-0.8em\TeX}}}

\setcopyright{acmcopyright}
\copyrightyear{2022}
\acmYear{2022}
\acmDOI{XXXXXXX.XXXXXXX}

\acmConference[MSR '22]{19th
International Conference on Mining Software Repositories (Registered
Reports)}{May 23--24, 2022}{Pittsburgh, Pennsylvania, United States}
\acmPrice{XX.XX}

\begin{document}

\title{Is Surprisal in Issue Trackers Actionable?}

\author{James Caddy}
\orcid{ 0000-0002-6254-8982 }
\affiliation{%
  \institution{University of Adelaide}
  \city{Adelaide}
  \country{Australia}
}
\email{james.caddy@adelaide.edu.au}

\author{Markus Wagner}
\orcid{ 0000-0002-3124-0061 }
\affiliation{%
  \institution{University of Adelaide}
  \city{Adelaide}
  \country{Australia}}
\email{markus.wagner@adelaide.edu.au}

\author{Christoph Treude}
\orcid{ 0000-0002-6919-2149 }
\affiliation{%
  \institution{University of Melbourne}
  \city{Melbourne}
  \country{Australia}}
\email{christoph.treude@unimelb.edu.au}

\author{Earl T. Barr}
\orcid{ 0000-0003-0771-7891 }
\affiliation{%
  \institution{University College London}
  \city{London}
  \country{United Kingdom}
}
\email{e.barr@ucl.ac.uk}

\author{Miltiadis Allamanis}
\orcid{ 0000-0002-5819-9900 }
\affiliation{%
  \institution{Microsoft Research}
  \city{Cambridge}
  \country{United Kingdom}
}
\email{miltos@allamanis.com}

\renewcommand{\shortauthors}{Caddy, et al.}

\begin{abstract}
\paragraph{Background}
From information theory, surprisal is a measurement of how unexpected an event is. Statistical language models provide a probabilistic approximation of natural languages, and because surprisal is constructed with the probability of an event occuring, it is therefore possible to determine the surprisal associated with English sentences. The issues and pull requests of software repository issue trackers give insight into the development process and likely contain the surprising events of this process.

\paragraph{Objective}
Prior works have identified that unusual events in software repositories are of interest to developers, and use simple code metrics-based methods for detecting them. In this study we will propose a new method for unusual event detection in software repositories using surprisal. With the ability to find surprising issues and pull requests, we intend to further analyse them to determine if they actually hold importance in a repository, or if they pose a significant challenge to address. If it is possible to find bad surprises early, or before they cause additional troubles, it is plausible that effort, cost and time will be saved as a result.

\paragraph{Method}
After extracting the issues and pull requests from 5000 of the most popular software repositories on GitHub, we will train a language model to represent these issues. We will measure their perceived importance in the repository, measure their resolution difficulty using several analogues, measure the surprisal of each, and finally generate inferential statistics to describe any correlations.
\end{abstract}



\maketitle
\setcounter{section}{0}
\section{Introduction}
\label{sec:introduction}
\noindent
Surprisal is a measure in information theory that can quantify how unexpected and thus how informative a word $w_t$ is given the words that precede it ($w_1,...,w_{t-1}$). A higher word surprisal value indicates that the current word is less expected given the context. In mathematical terms, surprisal is defined as the negative logarithm of the word's conditional probability of occurrence~\cite{armeni2017probabilistic}.

In this work, we propose to apply the surprisal measure to software engineering artefacts, motivated by many researchers arguing that software developers need to be aware of unusual or surprising events in their repositories, e.g., when summarizing project activity~\cite{treude2015summarizing}, notifying developers about unusual commits~\cite{leite2015uedashboard, goyal2018identifying}, and for the identification of malicious content~\cite{gonzalez2021anomalicious}. The basic intuition is that catching bad surprises early will save effort, cost, and time, since bugs cost significantly more to fix during implementation or testing than in earlier phases~\cite{dawson2010integrating}, and by extension, bugs cost more the longer they exist in a product after being reported and before being addressed.

Following recent work on applying natural language techniques to software engineering data~\cite{hindle2016naturalness}, in this work, we investigate whether the information-theoretic measure of surprisal is actionable when applied to software repositories. In this study we analyse issues and pull requests separately, but as a matter of convention, we will refer to these simply as issues going forward. In our method, the differences are negligible, but separately they may produce unique analyses. 

We investigate two scenarios involving surprisal: its effect on resolution difficulty and the perceived importance of surprising issues. 
Prior work~\cite{kavaler2017perceived} found that conformance to project-specific language norms reduces issue resolution time. Assuming that surprising issues do not conform to such norms, we investigate whether they are more difficult to resolve. Other work~\cite{mohamed2018predicting} has analysed what kind of issues are reopened, and specific issue metrics such as number of comments are shown to correlate. We conceptualise difficulty in terms of (1) reopened rate, (2) amount of discussion, and (3) resolution time.
Further, prior work established that developers want to be aware of unexpected issues in their repositories~\cite{treude2015summarizing}, and that high importance issues are more likely to be included in release notes~\cite{abebe2016empirical}. In this work, we investigate whether surprising issues are treated with higher importance. We conceptualise importance in terms of (1) mention in release notes, (2) first issues to be worked on after a break, (3) issues attracting GitHub reactions~\cite{borges2019beyond}, and (4) priority labels being assigned to issues.

\subsection{Motivating Examples}
\label{sec:motivatingexample}
Take for example, a software product sustainment team that receives a steady stream of issue reports which they must triage. Each of these issues have a cost increasing at a given rate, which itself might be accelerating. This cost may be in safety, budget, and/or reputation for example. It is in the interest of stakeholders that cost is minimised across the lifetime of the product.

In order to predict the cost of the new issues coming in, someone with experience needs to spend time comparing against previous experience, or applying metrics based methods. Then time needs to be spent addressing the issue for the cost to be eliminated.

If surprisal can be used to tell which issues are challenging or notable without human input, it is possible that efficiency can be gained in the resolution process. The triage process is better informed with notable issues that might need to be prioritised sooner. The cost to address an issue, weighed against its cost to the project, is better informed by the challenge that the issue presents. Challenging issues can have more people assigned to resolve it. Information about notable issues can be distributed to other developers so they know what mistakes to avoid in the future.

Another example is of a new developer just joining a project. In order to familiarise themselves with the history and most important changes or developments that have occurred, they would usually have to rely on release notes or a version history. In the case where the project either lacks release notes, or in the case where the release notes contain even the most minor changes, this can be overwhelming and of limited value to the new developer. A tool based on surprisal could extract the most important or notable changes from a mature project for this developer, or even tell someone who has been away from a project for a period what notable things have happened since they left.

\section{Background}
In Claude Shannon's seminal work on information theory he describes radio signals and their ability to communicate information. In doing so, he makes the first formal description of information entropy or information uncertainty.

\subsection{Uncertainty of an Event}
Take two events, represented by the symbols `A' and `B'. If we know that these events are equally likely to occur, we could say that we are equally uncertain which the next event will be. The information that we gain from observing an event can be represented by $I(x)=-\log_{2}{P(x)}$, where $P(x)$ is the probability of $x$ occurring. In this example, we can see that for either outcome, $P(A)=P(B)=0.5$ and so $I(A)=I(B)=1$. We gain exactly 1 bit of information, as there are two possible outcomes and either is as likely to occur.

If we take another example, and modify the probabilities of the outcomes such that $P(A)=0.0$ and $P(B)=1.0$, we can see that if we observe B, we gain exactly 0 bits of information from $I(B)=-\log_2{1}=0$. This is intuitive, since we already knew that the event would be B, there was no uncertainty.

Now we take an example where an event is extremely unlikely to happen. When $P(A)=0.999$ and $P(B)=0.001$, if we observe the event A, we gain $I(A)=-\log_{2}{0.999}\approx{}0.001$ bits of information. Since it was likely to happen, we do not receive much information, but still a little. It is then surprising when we observe the event B, since it is unlikely to happen. The information we gain from observing B, $I(B)=-log_{2}{0.001}\approx{}9.966$, is extreme in comparison.

\subsection{Statistical Language Models}
Shannon describes an approximation of the English language by utilising n-gram statistical language models. Initially he describes a crude unigram model that selects the next word in a sentence based on their relative frequencies in the English language. This has an obvious flaw; unigram models assume that each choice of word is independent from the last. Syntax and grammar demand more careful choice of words than pure randomness alone, so to account for this, greater order n-gram models are used.

Bigram models select the next word based on the previous word, and trigrams select based on the previous two. These better capture the structure of English, and while higher order models are possible, these tend to overfit the training data. The language model also becomes exponentially sparser as the order increases, which will in turn require exponentially more training data or risk severely underfitting the training data.

With a model that statistically represents the English language in n-grams, we can now determine how likely a word $w_t$ is, given the words that precede it ($w_1,...,w_{t-1}$).

\subsection{Probability Distributions}
To determine how closely an issue's description represents the whole corpus of descriptions, in an effort to see how surprising or not it is, we can use cross entropy. To understand cross entropy, it is important to discuss the underlying concept of entropy. Entropy can be quantified as the average number of symbols needed to represent an event from a distribution of events. Take for example, a distribution of events where:
\begin{align}
    \nonumber P(A)&=0.5,\\
    \nonumber P(B)&=0.25,\\
    \nonumber P(C)&=0.25
\end{align}
This distribution is biased, or skewed in favour of observing the event `A'. We can describe these events with their relative frequencies in a Shannon–Fano coding \cite{fano1949transmission}, where:
\begin{align}
    \nonumber A&=\{0\},\\
    \nonumber B&=\{01\},\\
    \nonumber C&=\{10\}
\end{align}
In this example, we can see that half of the time, we only require $1$ bit of information to represent any event. The other half of the time, we require $2$ bits. We can therefore say that the average number of bits required to represent an event from this distribution (its entropy) is $1.5$. Just as above, where an event that occurs often can be represented with less bits, distributions that are more biased, that yield one particular event more often, present less entropy. Shannon formalises this into the following equation, using $X$ as the support for random events $x$:
\begin{equation}
    H(P)=-\sum_{x\ \in\ X}P(x)\times\log(P(x))
    \label{eq:entropy}
\end{equation}

Cross entropy describes how many symbols on average are required to represent an event from one distribution in a coding optimised for another, if both have the same support. In our setting, it measures how many symbols are required to represent an issue's description, based on the distribution of words observed ($P_o$) in the issue, in the true distribution ($P_\mathit{tt}$) of words in all issues.

\begin{equation}
    H(P_o,P_\mathit{tt})=-\sum_{x\ \in\ X}P_{o}(x) \times \log(P_\mathit{tt}(x)) \label{eq:crossentropy}
\end{equation}

When $P_{o}=P_\mathit{tt}$, \Cref{eq:crossentropy} is equivalent to \Cref{eq:entropy}. Otherwise, when the observed distribution (the issue's use of words) differs from the distribution of the training set (the set of all issues), cross entropy increases. Higher values of cross entropy therefore imply an issue is more surprising.

There are potentially other surprisal metrics that provide different evaluations. More simple measures for instance, might take the minimum, maximum, or average surprisal of all words in an issue. Cross entropy is a very well-realised metric of model accuracy in the literature~\cite{niven2007combinatorial}, and it is for this reason that we are using it.

\section{Research Questions}
A number of research questions and hypotheses have been made to guide the investigation.
\begin{tcolorbox}[enhanced, frame hidden, sharp corners]
\begin{enumerate}
    \item[\textbf{RQ1}:]{How well does information theory's surprisal, as measured by a statistical language model, align with perceived surprisal?}
\end{enumerate}
\end{tcolorbox}

With RQ1, we hope to find how statistical language models compare to the human perception of surprisal, and also which factors of the language model influence its ability to measure the surprisal of an issue.

\begin{tcolorbox}[enhanced, frame hidden, sharp corners]
\begin{enumerate}
    \item[\textbf{RQ2}:]{Are surprising issues correlated with resolution difficulty?}
\end{enumerate}
\end{tcolorbox}

In RQ2, we define ``resolution difficulty'' as a combination of the following factors. Difficult to resolve issues are: reopened more often; attract more discussion prior to resolution; and take longer to be resolved, when compared to issues with little or no resolution difficulty. This is perhaps an incomplete list, but will serve as the basis for the report.

RQ2 can be formalised into the following hypotheses:
\begin{enumerate}
    \item[H\textsubscript{2.1}:]{Surprising issues are more likely to be reopened.}
    \item[H0\textsubscript{2.1}:]{There is no significant difference in how often surprising issues are reopened, compared to unsurprising issues.}
    \vspace{0.5em}
        
    \item[H\textsubscript{2.2}:]{Surprising issues attract more discussion. (Number of people involved)}
    \item[H0\textsubscript{2.2}:]{There is no significant difference in how much discussion surprising issues draw, compared to unsurprising issues.}
    \vspace{0.5em}
    
    \item[H\textsubscript{2.3}:]{Surprising issues attract more discussion. (Number of interactions)}
    \item[H0\textsubscript{2.3}:]{There is no significant difference in how much discussion surprising issues draw, compared to an unsurprising issue.}
    \vspace{0.5em}
    
    \item[H\textsubscript{2.4}:]{Surprising issues take longer to resolve.}
    \item[H0\textsubscript{2.4}:]{There is no significant difference in time to resolve surprising issues, compared to unsurprising issues.}
    \vspace{0.5em}
    
    \item[H\textsubscript{2.5}:]{Surprising issues are difficult, and difficult issues are best represented as some combination of reopen rate, amount of discussion, and time to resolve.}
    \item[H0\textsubscript{2.5}:]{Surprising issues are difficult, but difficulty is best represented as only one factor of reopen rate, amount of discussion, or time to resolve.}
\end{enumerate}

\begin{tcolorbox}[enhanced, frame hidden, sharp corners]
\begin{enumerate}
    \item[\textbf{RQ3}:]{Are surprising pull requests correlated with resolution difficulty?}
\end{enumerate}
\end{tcolorbox}

As in RQ2, RQ3 defines difficulty in the same way but for pull requests rather than issues. The purpose of RQ3 is to determine if the more structured and formal contents of pull requests are more suitable than issues for establishing a relationship between surprisal and difficulty. Pull requests offer a suitable alternative because they are intended to directly address a need that would typically be expressed in an issue. Additionally, merged pull requests are reviewed and thus are unlikely to be duplicated, describe what they are resolving, and what they address is far less likely to be misreported as a defect if it is not.

The formal hypotheses for RQ3 are formulated in a manner identical to the hypotheses of RQ2, since pull requests and issues are functionally identical in this context. In the interest of brevity, the full text of these hypotheses H\textsubscript{2.1} through H\textsubscript{2.5} has been omitted.

\begin{tcolorbox}[enhanced, frame hidden, sharp corners]
\begin{enumerate}
    \item[\textbf{RQ4}:]{Are surprising issues correlated with perceived importance?}
\end{enumerate}
\end{tcolorbox}

In RQ4, we define ``perceived importance'' as a combination of the following factors. Important issues are: mentioned in release notes; addressed soon after periods of breaks; attract more GitHub reactions; and have high-priority labels added. Once again, this is perhaps an incomplete list, but will serve as the basis for the report.

RQ4 can be formalised into the following hypotheses:
\begin{enumerate}
    \item[H\textsubscript{4.1}:]{Surprising issues are more likely to be mentioned in release notes.}
    \item[H0\textsubscript{4.1}:]{There is no significant difference in how often surprising issues are mentioned in release notes, compared to unsurprising issues.}
    \vspace{0.5em}
    
    \item[H\textsubscript{4.2}:]{Surprising issues are addressed with priority over unsurprising issues after a hiatus.}
    \item[H0\textsubscript{4.2}:]{There is no significant difference in the time addressing post-hiatus surprising issues, compared to unsurprising issues.}
    \vspace{0.5em}
    
    \item[H\textsubscript{4.3}:]{Surprising issues attract more GitHub reactions.}
    \item[H0\textsubscript{4.3}:]{There is no significant difference in how many reactions a surprising issue receives, compared to unsurprising issues.}
    \vspace{0.5em}
    
    \item[H\textsubscript{4.4}:]{Surprising issues are more often labelled as high-priority issues.}
    \item[H0\textsubscript{4.4}:]{There is no significant difference in what priority surprising issues are labelled, compared to unsurprising issues.}
    \vspace{0.5em}
\end{enumerate}

In H\textsubscript{3.2}, we define a hiatus as the top 25\% longest times between issue resolutions per contributor for a particular repository. Priority in this case is the order in which issues are worked on after a hiatus.

\begin{tcolorbox}[enhanced, frame hidden, sharp corners]
\begin{enumerate}
    \item[\textbf{RQ5}:]{Are surprising pull requests correlated with perceived importance?}
\end{enumerate}
\end{tcolorbox}

As before with RQ2 and RQ3, RQ4 focuses on issues, and RQ5 will focus on their pull request counterparts. The full text of  hypotheses H\textsubscript{4.1} through H\textsubscript{4.4} has been omitted for brevity.

\vspace{1em}

\section{Variables}
A summary of the variables involved for all the research questions can be found in \Cref{tab:variables}.

\noindent\textit{Predictor Variable}:
\begin{itemize}
    \item \textbf{Surprisal.} How surprising the content of the issue is to the language model. This is calculated as the cross entropy of the issue text and the corpus of issues.
\end{itemize}

\noindent\textit{Response Variables}:
\begin{itemize}
    \item \textbf{Reopenings.} How many times the issue has been reopened. An issue can be labelled as closed and then reopened for numerous reasons, just as a pull request can be merged and then reopened. This usually indicates a regression of functionality, reoccuring bug, or unsuccessful fix~\cite{jiang2019characteristics}, all indications that the issue has additional complexity associated with it.
    \item \textbf{Participants.} How many individual participants have interacted with the issue. Every event that takes place on an issue has an actor associated with it. This actor represents somebody interacting with the issue. If a particular issue involves multiple assignees for example, it may be a sign that additional expertise is needed to resolve it. It could also mean that it affects a lot of people but is not necessarily more difficult. Kavaler \textit{et. al.}~\cite{kavaler2017perceived} show that there is a significant increase in issue resolution time with an increased number of unique participants.
    \item \textbf{Interactions.} How many interactions have been made with the issue, including comments, mentions, taggings, assignments and state changes. A full list of events is available in the GitHub Issue API documentation~\cite{githubevents}. Some of these interactions are considered a normal part of the resolution process, although we expect to see more interactions if the issue reveals hidden complexity over time. Kavaler \textit{et. al.}~\cite{kavaler2017perceived} also show there is an increase in issue resolution time corresponding to the number of comments made.
    \item \textbf{Open State Duration.} How long the issue has been unresolved; from first submission to last closure, or if it has not been closed, the time of analysis. While some issues may not be difficult but especially time consuming, we expect to see longer resolution times for issues that are difficult to diagnose or replicate.
    \item \textbf{Mentions in Release Notes.} How many times the issue has been mentioned within release notes on GitHub. Some repositories make no use of GitHub's Releases feature, so only the repositories that do, and that mention issues at all in them will be considered. Highly important issues are more likely to be included in release notes~\cite{abebe2016empirical}.
    \item \textbf{Order of Address.} After lengthy breaks of development, whether independently taken or due to holidays, it is likely that the most pressing of issues in the backlog are chosen for immediate resolution. Commits after extended breaks have been described as interesting, in a previous paper~\cite{treude2015summarizing}. Each contributor has a time between addressing issues, so taking the top 25\% of these breaks, and then ordering the issues that they worked on afterwards gives us an indication of what importance that author places on each issue.
    \item \textbf{Reactions.} How many reactions have been made on the issue. Reactions give a quick way for users to interact with an issue. For example, users can express joy that a particular issue is closed, or frustration if it disrupts them personally, through reactions. This can be seen as a community rating of importance, rather than that of the maintainers~\cite{borges2019beyond}.
    \item \textbf{Labelling.} Many repositories use the GitHub issue labelling system to triage incoming bug reports and feature requests. Issues are sorted by maintainers into priorities and labelled as such, from low priority to high priority~\cite{xie2021mula},~\S3.B. These labels can be seen as the maintainer's rating of importance.
\end{itemize}
\vspace{1em}

\renewcommand{\arraystretch}{1.2}
\begin{table*}
  \caption{Variables}
  \label{tab:variables}
  \rowcolors{2}{gray!25}{}
  \begin{tabular}{llp{5.5cm}lp{5cm}}
    \toprule
\textbf{Variable} & \textbf{Hypotheses} & \textbf{Description} & \textbf{Measure} & \textbf{Operationalisation} \\
\midrule
Surprisal & \pbox[t]{3cm}{Predictor for \\all hypotheses} & How surprising an issue is to the statistical language model. & Ratio & Cross entropy of issue, obtained with probability from SLM trained on corpus of all issues.\\
Reopenings & \pbox[t]{3cm}{Response for \\H\textsubscript{1.1}, H\textsubscript{2.1}} & After an issue has been labelled as closed or resolved, it can be reopened either due to a fix being unsuccessful, a regression, or reoccurring bug. & Ratio & GitHub Issue API's ``reopened'' event.\\
Participants & \pbox[t]{3cm}{Response for \\H\textsubscript{1.2}, H\textsubscript{2.2}} & Number of unique individuals involved with the issue. Each event (as described by the GitHub API~\cite{githubevents}) associated with an issue has an actor that initiates it, whether that event is a comment, or a state update. & Ratio & GitHub Issue API's event ``actor'' for each event associated with an issue.\\
Interactions & \pbox[t]{3cm}{Response for \\H\textsubscript{1.3}, H\textsubscript{2.3}} & Number of events associated with the issue. & Ratio & Count of events, as described by the GitHub Issue API~\cite{githubevents}.\\
\pbox[t]{2cm}{Open State \\Duration} & \pbox[t]{3cm}{Response for \\H\textsubscript{1.4}, H\textsubscript{2.4}} & Length of time between the issue's creation, and it being resolved for the final time. & Ratio & Difference between GitHub Issue API's ``created\_at'' value for the issue and final ``closed'' or ``merged'' event.\\
Mentions & \pbox[t]{3cm}{Response for \\H\textsubscript{3.1}, H\textsubscript{4.1}} & Number of mentions within a repository's release notes. & Ratio & Scrape for issue numbers through GitHub's Releases API.\\
\pbox[t]{2cm}{Order of \\Address} & \pbox[t]{3cm}{Response for \\H\textsubscript{3.2}, H\textsubscript{4.2}} & Issue order after top 25\% of contributor's inter-issue resolution times. & Interval & Issues assigned to a contributor are retrieved through the GitHub Issues API.\\
Reactions & \pbox[t]{3cm}{Response for \\H\textsubscript{3.3}, H\textsubscript{4.3}} & Number of reactions on a particular issue. & Ratio & Count of reactions for an issue, from GitHub's Reactions API.\\
Labelling & \pbox[t]{3cm}{Response for \\H\textsubscript{3.4}, H\textsubscript{4.4}} & Assigned priority or importance label of a particular issue. & Interval & Labels are extracted via the GitHub Issues API, and then normalised (per repository) to a 3-degree scale, `low-importance', 'regular-importance', `high-importance'.\\
    \bottomrule
  \end{tabular}
\end{table*}

\section{Data Sets}
In this section we present the data sources that we will use, and how we will use them.
\subsection{Sources}
For this study, open-source software stored on GitHub serves as our primary and sole source for software issues. Unfortunately, GitHub imposes a restrictive limit to how many interactions with its API a user can make. Typically this is 5000 calls per hour~\cite{githubratelimit}, and so we take the top 5000 most `starred' repositories as our data set. Stars represent a user liking a repository, and therefore indicate popular repositories, more likely to have high numbers of issues due to increased testing and feature requests --- a result of more users~\cite{borges2016understanding}.

This supposes that GitHub is used in the same way by the maintainers of those 5000 repositories, which is not the case. Many of these repositories do not make use of the Issues feature; many of these repositories are not software development related and possess little to no code; and other repositories are simply a mirror of a repository developed and hosted elsewhere~\cite{kalliamvakou2014promises}. Additionally, we wish to limit the scope to English language repositories. To find the repositories that fit these conditions, we plan to use G-Repo~\cite{romano2021grepo} as a means to filter out non-software repositories, repositories making little use of the Issues feature (less than 1000 issues), and non-English repositories.

\subsection{Language Model Transfer}
The statistical language model (SLM) used to determine the surprisal of an issue, is intended to represent a probabilistic model of what the content of an issue looks like. Software issues are typically written in such a way that requires domain-specific knowledge of terminologies and jargon, and are in most cases very specific to the project they reference. As a result, a more specific SLM, trained on a software-based corpus would see some improvement in accurately modelling the software language used, compared to a more general pre-trained model~\cite{wang2021well},~\S5.3. For this reason, it was decided that a bespoke language model will be trained for the task.

\subsection{Pre-processing}
Issues are composed of a title and description. Both of these elements have the possibility of individually containing pertinent information, and in some examples do not describe the information the other holds. It is for this reason that during the pre-processing stage, we will prepend the issue description with the issue title. The SLM will then be trained on these title-description combinations.

Before training a model on the issue text, we will clean the data with the following pre-processing steps:
\begin{enumerate}
    \item{HTML Void Elements~\cite{htmlvoidelements} are translated into special tokens, e.g., \texttt{<br>} becomes \texttt{[BR]}. \\(Also code blocks, see details following.)}
    \item{Other HTML elements are replaced with their contents, e.g., a list element becomes a simple string of its content.}
    \item{Text undergoes normalisation of its Unicode forms. Canonical Composition (NFC) is used in accordance with the Character Model standard proposed by W3C~\cite{w3ccharmodnorm}.}
    \item{Punctuation and symbols are removed on word boundaries, and when isolated.}
    \item{Stop words are removed, and remaining words are stemmed. \\(See details following.)}
\end{enumerate}

As singular code tokens --- variable names and the like --- may provide important contextual information across multiple issues, they will be preserved in the training data. Code blocks on the other hand risk introducing too many globally-unique tokens into the model without introducing useful and actionable information to the description. The reason is that the syntax of code is entirely disparate with that of English for example, and so code blocks will be replaced with a special token \texttt{[CODE]} where they are demarcated with the \texttt{<pre>} and \texttt{<code>} HTML tags (as is typical on GitHub).

It is possible that the quality of the language model could be improved by further transforming the text in a final step. In many natural language processing applications, stop word removal is conducted to remove low-information tokens, thereby increasing the average entropy. In other applications, lemmatisation or stemming is used to reduce the occurrence of high-information tokens. Concepts are learned better if you can reduce ``fishing'' and ``fishlike'' to ``fish''. However we cannot know for certain what using either will achieve when modelling surprisal, and a preliminary investigation will be conducted to see if it indeed increases the accuracy of the model. The choice of algorithms is still to be determined, and will be revisited in the final report.

\subsubsection{Priority Labelling}
\label{sec:prioritylabelling}
Some GitHub repositories use the labelling system to assign priority grades or importance to issues in their triage process. These labels are user text input and can represent these grades in a number of different ways. For example, one repository may use the labels `Low Priority', and `High Priority', where another may use the labels `P1' through `P5'.

In order to process these priorities, it is necessary for manual classification to normalise these different ranges. For the purpose of this experiment, priority is distributed among the three degrees `low-importance', `regular-importance', and `high-importance'. Tie-breaking is settled by ruling in favour of a higher importance, using the previous examples for example; `low-importance' would take `P1' and `Low Priority'; `regular-importance' would take `P2' and `P3'; and `high-importance' would take `P4', `P5', and `High Priority'.

To ensure that these have been correctly classified, we propose giving two researchers a list of 400 randomly selected labels from the population of approximately 9000 different labels and asking them to classify them using the method above, or classify them as `unrelated' to priority or importance. The agreement between the researchers is then calculated using Cohen's kappa, and in the event that a consensus value of 0.7 is reached, we consider the task sufficiently unambiguous, and a single researcher can then be trusted to accurately represent the rest of the $\sim9000$ labels. If not, a more involved process needing agreement by more researchers is required to classify the labels.

\section{Execution Plan}
\label{sec:executionplan}
In this section we present the method we plan to use for the study. It is important that all the decisions made during the experiment will be recorded, along with intermediate outputs. For example: a record will be made listing the 5000 most starred repositories on GitHub at the time of the experiment; after repositories are removed when failing to meet selection criteria, those will also be made into a record; and decisions such as how the total issue count was calculated will be made into a record too. All the code used will be made available in a GitHub repository for the final report.
\subsection{Method}
To test all hypotheses the following method is proposed:

\begin{enumerate}
    \item Using the GitHub API, query the 5000 most starred repositories on GitHub.
    \item Query the number of issues. If any repository has less than 1000 issues, remove it from the pool.
    \item Extract all GitHub issues from each repository.
    \item Train SLM on entire issue description corpus.
    \item For each repository:
    \begin{enumerate}
        \item Determine if it uses priority/importance labels, and normalise those labels into three degrees of importance. Ties broken in favour of being more important.
        \item For each issue:
        \begin{enumerate}
            \item Determine and record surprisal of issue.
            \item Record how many re-openings have occurred.
            \item Record how many unique participants have interacted with the issue.
            \item Record how many interactions have been made with the issue.
            \item Calculate and record open duration for the issue.
            \item Record number of reactions to the issue.
            \item Record normalised importance label, if any.
        \end{enumerate}
        \item Parse all release notes for the repository, incrementing the mentions of a particular issue when it appears.
        \item For each contributor:
        \begin{enumerate}
            \item Determine top 25\% of inter-issue resolution times.
            \item Split all issues assigned to the contributor into bands following a break of at least the 25th percentile's length.
            \item Order the issues by oldest assigned time.
            \item Record the ordinal position of each issue in its inter-break band.
        \end{enumerate}
    \end{enumerate}
    \item Split issues from GitHub into pull requests and simple issues.
    \item Generate descriptive statistics for each repository according to analysis plan.
    \item Generate inferential statistics across entire issue data set according to analysis plan.
\end{enumerate}

\begin{figure}
\centering
\begin{tikzpicture}[node distance=1cm]
    \node (start)   [startstop] {Start};
    \node (io-github)   [io, below of=start]    {Query GitHub API};
    \node (proc-pre)    [process, below of=io-github]   {Pre-processing};
    \node (proc-calc)   [process, below of=proc-pre]   {Calculate Difficulties};
    \node (proc-calc2)  [process, below of=proc-calc]   {Calculate Importance};
    \node (proc-train)  [process, right of=proc-calc, xshift=2cm]   {Train SLM};
    \node (proc-sel)    [process, below of=proc-calc2]  {Select Test Issue};
    \node (io-stat)   [io, below of=proc-sel]    {Generate Statistics};
    \node (end) [startstop, below of=io-stat] {End};
    \draw [arrow]   (start) -- (io-github);
    \draw [arrow]   (io-github) -- (proc-pre);
    \draw [arrow]   (proc-pre) -| (proc-train);
    \draw [arrow]   (proc-train) |- (proc-sel);
    \draw [arrow]   (proc-pre) -- (proc-calc);
    \draw [arrow]   (proc-calc) -- (proc-calc2);
    \draw [arrow]   (proc-calc2) -- (proc-sel);
    \draw [arrow]   (proc-sel) -- (io-stat);
    \draw [arrow]   (io-stat) -- (end);
\end{tikzpicture}
\caption{Method Flowchart}
\end{figure}
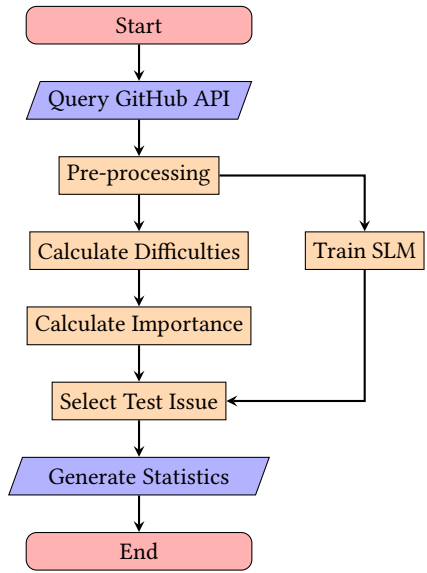

\subsection{Statistical Language Modelling}
N-gram models saw some popularity during the 2000s and 2010s, but gradually saw declining popularity due to their perceived contextual fragility~\cite{rosenfeld2000decades}(2000), and competitive neural language models being developed~\cite{mikolov2012statistical}(2012). Despite this, n-gram models still perform admirably in Natural Language Processing applications. For our SLM, we use a trigram model using word tokens (shingles). We chose this due to its simplicity, familiarity, popularity, and reasonably effective results~\cite{chen1999empirical}. More modern statistical approaches typically use a n-gram model utilising a backoff strategy to deal with unknown tokens.

We train the SLM using the entire corpus of issues. This gives us the advantage of never coming across an unknown token, as all the tokens that we generate surprisal values for are contained in the training set. Any unknown, in this case `unique', tokens would otherwise be infinitely surprising, which might not accurately represent how common it is to appear in say a larger sample size of repositories and issues.

The loss of generality realised by the lack of testing set is insignificant when compared to the small sample size of 5000. For this study, we are more concerned with the formulation of surprisal, and the two correlations applying to the repositories we have selected. Transferability can be further investigated after this proof of concept, if successful. Conversely, because singular repositories have a small sample size of issues, the specificity gained by training only on a single repository introduces massive sparsity in the model. This is another reason we include all issues into the training set.

\subsection{Model Smoothing}
Despite the lack of unknown tokens, we still apply a smoothing algorithm to obtain a more uniform, less sparse, and more representative model of the larger corpus of issues outside of the repositories that we have selected. The modified Kneser-Ney algorithm described by Chen and Goodman~\cite{chen1999empirical},~\S3, was chosen due to its excellent smoothing performance.

\subsection{Model Improvement}
\label{sec:modelimprovement}
While trigrams were chosen for the reasons previously mentioned, it is worth understanding how n-gram order affects the language model's calculation of surprisal. It is also worth understanding how the use of a testing set from the training data affects this calculation of surprisal.

These two factors contribute to the SLM and reveal important information regarding RQ1. In order to measure the affects of these factors, we propose the following additional experiment which compares the SLM performance against a manual classification:

\begin{enumerate}
    \item A repository of relatively few issues is chosen as a model.
    \item Take a representative sample of issues from the repository.
    \item Two researchers are asked to rate the surprisal in each issue, after reading all issues thoroughly.
    \begin{enumerate}
        \item Issues are rated on a Likert scale.
        \item A value of 1 means that the description of an issue is not surprising.
        \item A value of 5 means that the issue contains almost completely unique information.
        \item Surprisal judgements should be based on factors such as topic, formatting, length, and word usage.
    \end{enumerate}
    \item Choose the SLM training set:
    \begin{enumerate}
        \item Entire issue descriptions corpus.
        \item Entire issue descriptions corpus, with the chosen repository absent.
        \item Entire issue descriptions corpus, with each issue removed individually.
    \end{enumerate}
    \item Choose n-gram order, repeating from 1-grams to 10-grams.
    \item Measure and record surprisal of each issue (the same individually removed from the training set if that type of training set is chosen).
    \item Generate agreement statistics between each model and the manual classification.
\end{enumerate}

\section{Analysis Plan}
\label{sec:analysisplan}
This is a correlational study of independent random variables, the following is a design of the quantitative analysis that will be undertaken. All calculations will be performed using a statistics package.

\paragraph{Model Analysis}
In \Cref{sec:modelimprovement} we propose a method with which the surprisal of a repository-representative sample of issues is judged manually by two researchers. The surprisal of the issues is then measured using the surprisal calculated by the SLM. We propose using Cohen's kappa to measure the agreement between the two researchers, and report how contentious this task is. We also propose using a Kendall rank correlation to measure the agreement between the SLM- and manually-computed surprisal values. In the event that the agreement is statistically significant, it can be said that SLMs can calculate surprisal to a similar degree as a human participant can, and are suitable for the following analyses. A qualitative analysis of the issues that have disagreement in the surprisal ratings is conducted. Both the case of inter-researcher disagreement and researcher-SLM disagreement will be analysed in an effort to understand what caused this disagreement.

We expect that a language model using the entire corpus of issue descriptions and trigrams sufficiently agrees with the researcher's judgements. However, we will run the deeper analysis using different training sets and n-gram orders to confirm those choices are indeed correct. A combination of each choice will maximise the agreement with the manual computation, and this combination will be used in the analyses going forward. In the event that even the combination that maximises the agreement is still not statistically significant, a more thorough inspection of the factors influencing the SLM is required.

\paragraph{Descriptive Statistics}
We will present descriptive statistics of the predictor and response variables in a summary of the complete data set. Sample size, mean, standard deviation, maximum, and minimum values for each variable will be included

\paragraph{Inferential Statistics}
Before presenting a correlation and regression analysis using a linear regression, we graph the relationship between the surprisal and each response variable for the issues in a repository. We will then test each hypothesis separately with its corresponding variable, and look for statistical significance. 

A Shapiro-Wilk test is used to determine if the data shows normality \cite{shapiro1965analysis}. If the data is normally distributed, we can choose to use the additional descriptive power of a Pearson correlation. If not, a Spearman correlation can be used instead \cite{spearman1987proof},~\S1.2.a).

In order to test H\textsubscript{1.5} and H\textsubscript{2.5}, we will perform a multiple linear regression, this time including one statistically significant measure of difficulty into the null model. If the F-test is \textbf{not} statistically significant, we can accept the alternate hypothesis. 

This test shows if a combination of our chosen difficulty factors is better in representing difficulty as a whole, in comparison to the measure added to the null model on its own. When performing this second regression, we expect to see multicollinearity, a high Variance Inflation Factor for these measures \cite{kutner2004applied},~\S10.5, as our belief is that they represent the same variable: difficulty. 

\paragraph{Interpretation of Results.}
We will present our interpretation of our findings, a discussion on any assumptions discovered, limitations, threats to validity, and future research.


\bibliographystyle{ACM-Reference-Format}
\bibliography{bib}


\end{document}